\documentclass{article}

\usepackage{microtype}
\usepackage{graphicx}
\usepackage{float}
\usepackage{subcaption}
\usepackage{caption}
\usepackage{booktabs}
\usepackage{multirow}
\usepackage{array}
\usepackage{siunitx}
\usepackage{xcolor}
\usepackage{colortbl}
\usepackage{enumitem}
\usepackage[normalem]{ulem}

\usepackage[table]{xcolor}
\usepackage{tabularx}
\usepackage{booktabs}
\usepackage{array}

\usepackage[T1]{fontenc}
\usepackage{lmodern}
\usepackage{tabularx}

\definecolor{gOneAcc}{HTML}{A04968}   \definecolor{gOneBg}{HTML}{EFE7EC}   
\definecolor{gOneFg}{HTML}{6B4D5D}
\definecolor{gTwoAcc}{HTML}{6B9456}   \definecolor{gTwoBg}{HTML}{E8EDE4}   
\definecolor{gTwoFg}{HTML}{5A6B53}
\definecolor{gThreeAcc}{HTML}{4A7FA8} \definecolor{gThreeBg}{HTML}{E6ECF0} 
\definecolor{gThreeFg}{HTML}{4F6675}
\definecolor{gFourAcc}{HTML}{B8884A}  \definecolor{gFourBg}{HTML}{F1EBE1}  
\definecolor{gFourFg}{HTML}{7A6A4F}
\definecolor{headfg}{HTML}{3D3D3D}

\definecolor{headfg}{HTML}{4A4A4A}   
\definecolor{rulegray}{HTML}{D8D3C6} 
\newcommand{\labelstyle}[1]{\textsc{\small #1}}

\newcommand{\grouprow}[4]{%
  \rowcolor{#1}[0pt][0pt]\multicolumn{3}{@{}l@{}}{%
    \color{#2}\labelstyle{#3}%
    \ifx\\#4\\\else\ \textnormal{\color{#2}\small #4}\fi%
  }\\[-0.5ex]
}

\newcommand{\feat}[1]{\textcolor{#1}}

\newlength{\tabwid}

\usepackage{hyperref}
\usepackage{multirow}
\usepackage{svg}

\usepackage[accepted]{icml2026}

\usepackage{amsmath}
\usepackage{amssymb}
\usepackage{mathtools}
\usepackage{amsthm}

\usepackage{tabularx}

\usepackage{xcolor}

\definecolor{posgreen}{HTML}{0B6E4F}
\definecolor{negred}{HTML}{B00020}

\newcommand{\gain}[1]{\textcolor{posgreen}{$\blacktriangle$#1}}
\newcommand{\loss}[1]{\textcolor{negred}{$\blacktriangledown$#1}}

\usepackage[capitalize,noabbrev]{cleveref}

\theoremstyle{plain}

\theoremstyle{definition}

\theoremstyle{remark}

\usepackage[textsize=tiny]{todonotes}

\icmltitlerunning{Submission and Formatting Instructions for ICML 2026}

\begin{document}

\twocolumn[
  \icmltitle{Reading Calibrated Uncertainty from Language Model Trajectories}



  \icmlsetsymbol{equal}{*}

  \begin{icmlauthorlist}
    \icmlauthor{Aliai Eusebi}{ucl}
    \icmlauthor{Alexander Herzog}{ucl,core}
    \icmlauthor{Xiaoyu Liang}{ucl}
    \icmlauthor{Marie Vasek}{ucl}
    \icmlauthor{Enrico Mariconti}{ucl}
    \icmlauthor{Lorenzo Cavallaro}{ucl}
\end{icmlauthorlist}
\icmlaffiliation{ucl}{University College London, London, United Kingdom}
\icmlaffiliation{core}{CORE64, London, United Kingdom}
\icmlcorrespondingauthor{Aliai Eusebi}{aliai.eusebi.16@ucl.ac.uk}
\icmlcorrespondingauthor{Alexander Herzog}{alexander.herzog.23@ucl.ac.uk}

  \icmlkeywords{Machine Learning, ICML}

  \vskip 0.3in
]



\printAffiliationsAndNotice{}  

\begin{abstract}
The maximum softmax probability (MSP) represents a default approach when evaluating uncertainty quantification for language model generation with structured output. Although cheap, it is often miscalibrated. Methods that probe the model's internal activations feed raw hidden states into opaque classifiers, reading activations as static snapshots and leaving implicit the layer-wise trajectory by which a representation is formed. Yet, similar endpoints can arise from very different paths, and how evidence accumulates, reinforces, or reverses across depth might reveal uncertainty that final probabilities obscure. We extract eleven scale-invariant geometric features, tracing the cumulative path of per-layer MLP updates, and feed them to a sparse linear probe. The probe outperforms MSP under selective abstention, with gains scaling with baseline miscalibration up to 21 AURC points. Because every feature has a closed-form geometric meaning, the probe's coefficients trace  \emph{how} and \emph{where} along depth errors take shape -- which layers commit prematurely, which contradict the running state, where trajectories drift away from their endpoint.
\end{abstract}

\section{Introduction}                                           


A trustworthy language model should know when it might be wrong. Calibrated uncertainty quantification (UQ) captures this requirement: a model’s confidence should reflect its likelihood of being correct. Two equally accurate models can nevertheless differ substantially in reliability. If a model assigns low confidence to most of its errors, uncertain predictions can be deferred, abstained from, or escalated for human review. By contrast, when confidence is poorly aligned with correctness, high-confidence errors become indistinguishable from reliable predictions. In clinical triage, the first model would set its uncertain cases aside for a clinician to review; the second would queue a life-threatening case behind a routine one, both stamped high-risk with equally strong confidence. Calibrated uncertainty therefore serves a triage function, enabling uncertainty to guide which predictions warrant intervention or verification.

The default approach to UQ in discrete choice settings is the Maximum Softmax Probability (MSP) \cite{vashurin2025benchmarking, dakhmouche2025can}, which uses the predicted token’s softmax probability as a confidence score. MSP incurs no additional computational cost and is often surprisingly competitive. However, it may inherit the well-known pathology of miscalibration: confidence scores that fail to reflect the true likelihood of correctness, often remaining high even when the prediction is wrong \cite{on_the_calibration_of_modern_neural_networks}. A parallel line of work reads from the model's activations directly, on the premise that the internal computation leading to a generation carries information about its truthfulness that the output distribution alone does not. \citet{azaria2023internal} demonstrated that a simple classifier trained on hidden activations can predict whether an LLM's answer is truthful, and a growing body of work has since traced the contours of this “geometry of truth" \cite{li2023inference, marks2023geometry, dakhmouche2025can, liu2024uncertainty, beigi2024internalinspector, azizian2025geometries}.


Yet across this literature, activations are typically read as static snapshots -- a hidden state extracted from one layer, or averaged across layers, and analyzed for the information it contains. This discards the layer-wise trajectory by which the representation is formed. A final hidden state is the endpoint of a path through representation space, and similar endpoints may arise from qualitatively different trajectories. Some representations may develop steadily, others may emerge late, fluctuate, or be partially reversed before settling. These trajectories are not incidental to the prediction as they encode how evidence is accumulated, reinforced, attenuated, or revised across depth. We therefore read activations not only as frozen states, but as representational trajectories; we show that their geometry reveals uncertainty that final probabilities alone may obscure.



We calibrate by tracing the answer-position residual stream as the cumulative path induced by per-layer MLP write-vectors during the forward pass. We summarize this trajectory by computing scale-invariant geometric features and feed them to a sparse linear probe. Confidence-calibration is evaluated under selective abstention via AURC \cite{geifman2018bias}. This leads to an interpretable UQ method that goes beyond single-point evaluation (MSP) while avoiding the opacity of dense probes on raw activations.

To summarize, our main contributions are as follows:

\begin{enumerate}
    \item We propose a compact set of geometric features that describe how representations evolve across network-depth, and feed them to a sparse linear probe that outperforms MSP under selective abstention, with gains scaling with the baseline's miscalibration.
    \item We consider end to end interpretability of the probe: each feature has a closed-form geometric meaning, and its coefficients reveal not only \emph{whether} the model is likely to err but \emph{also how}, which layers commit prematurely, which contradict the running state, and where trajectories drift away from their endpoint. We show that correct and wrong predictions sharing the same MSP score leave different trajectory signatures, exposing information otherwise flattened by the output distribution.
    \item We perform a comprehensive set of empirical experiments across 9 instruction-tuned LLMs from three model families (Qwen, Llama, DeepSeek) spanning 3B to 72B parameters on five representative natural language processing tasks.
\end{enumerate}

\begin{table*}[t]
\centering
\setlength{\tabwid}{\linewidth}
\small
\setlength{\tabcolsep}{14pt}
\renewcommand{\arraystretch}{1.4}
\setlength{\aboverulesep}{2pt}
\setlength{\belowrulesep}{2pt}
\arrayrulecolor{rulegray}
\newcolumntype{F}{>{\hsize=0.85\hsize\centering\arraybackslash}X}
\newcolumntype{I}{>{\hsize=1.15\hsize\raggedright\arraybackslash}X}
\begin{tabularx}{\tabwid}{@{} l F I @{}}
\specialrule{0.9pt}{0pt}{0pt}
\textcolor{headfg}{\labelstyle{Feature}} &
\textcolor{headfg}{\labelstyle{Formula}} &
\textcolor{headfg}{\labelstyle{Interpretation}} \\
\specialrule{0.4pt}{0pt}{4pt}
\grouprow{gOneBg}{gOneFg}{Depth distribution}{}
\feat{gOneFg}{Relative update magnitude}    & $\|m_\ell\| / \mu$                                           & Per-layer update size vs.\ the mean \\
\feat{gOneFg}{Cumulative path fraction}     & $T^{-1} \sum_{k \leq \ell} \|m_k\|$                          & Front- vs.\ back-loading of computation \\
\grouprow{gTwoBg}{gTwoFg}{Local trajectory shape}{}
\feat{gTwoFg}{Consecutive cosine}            & $\cos(m_{\ell-1},\, m_\ell)$                                 & Directional consistency of updates \\
\feat{gTwoFg}{Curvature}                     & $1 - \cos(m_{\ell-1},\, m_\ell)$                             & Abrupt changes of direction \\
\feat{gTwoFg}{Update-state alignment}        & $\cos(m_\ell,\, s_{\ell-1})$                                 & Reinforcement vs.\ contradiction with prior state \\
\grouprow{gThreeBg}{gThreeFg}{Relation to endpoint}{$\hat{s}_L=s_L/\|s_L\|$}
\feat{gThreeFg}{Direction to final}          & $\cos(s_\ell,\, s_L)$                                        & State convergence to endpoint \\
\feat{gThreeFg}{Update to final}             & $\cos(m_\ell,\, s_L)$                                        & Update alignment with endpoint direction \\
\feat{gThreeFg}{Signed final support\textsuperscript{\textdagger}} & $\langle m_\ell,\, \hat{s}_L\rangle / \|m_\ell\|$      & Cosine alignment, magnitude-aware framing \\
\feat{gThreeFg}{Contradictory support}       & $\max\bigl(0,\, -\langle m_\ell,\, \hat{s}_L\rangle / \|m_\ell\|\bigr)$ & Magnitude of opposition to endpoint \\
\feat{gThreeFg}{Orthogonal mass fraction}    & $\sqrt{1 - \cos^2(m_\ell,\, s_L)}$                           & Fraction of update orthogonal to endpoint \\
\grouprow{gFourBg}{gFourFg}{Efficiency}{}
\feat{gFourFg}{Cumulative coherence}         & $\|s_\ell\| / \sum_{k \leq \ell} \|m_k\|$                    & Net displacement vs.\ path length \\
\specialrule{0.9pt}{0pt}{0pt}
\end{tabularx}
\caption{Eleven per-layer trajectory features  grouped by what they measure. Symbols are defined in the text. Features involving $m_{\ell-1}$ or $s_{\ell-1}$ are undefined at $\ell = 1$ and assigned conventional values ($1$ for \emph{Consecutive cosine}, $0$ for \emph{Curvature} and \emph{Update-state alignment}). \textsuperscript{\textdagger} \emph{Signed final support} is numerically identical to \emph{Update to final}; both rows are retained because they correspond to distinct geometric formulations (the L1 probe is invariant to this).}
\label{tab:features}
\end{table*}

\section{Background}

\subsection{Uncertainty Quantification}

Predictive uncertainty is commonly decomposed into two components: epistemic uncertainty, which stems from limited data or model misspecification and in principle reducible, and aleatoric uncertainty, which reflects the intrinsic stochasticity of the data-generating process and therefore irreducible \cite{kendall2017uncertainties}. A related dichotomy has emerged in the LLM literature: factual uncertainty pertains to the correctness of generated content with respect to ground-truth knowledge, whereas semantic uncertainty arises from the multiplicity of valid continuations admitted by a prompt; the former is epistemic in nature, the latter aleatoric \cite{liu2025uncertainty}. As factual uncertainty is the primary concern in automated decision-making \cite{dakhmouche2025can}, we restrict our analysis to this component. We employ multiple-choice questions, which act as a noise-reduction mechanism that isolates factual gaps by neutralizing the semantic uncertainty found in open-ended text \cite{li2026semantic}.

UQ methods aim to associate each prediction with a scalar score indicative of its correctness, enabling downstream decisions such as abstention, deferral, or selective generation. Existing approaches for LLM uncertainty estimation fall into three groups that differ primarily in computational cost. Single-sample methods derive uncertainty from a single forward pass, using signals such as maximum token log-probability \cite{manakul2023selfcheckgpt}, perplexity \cite{margatina2023active}, and entropy \cite{kadavath2022language, kuhn2023semantic}. Multi-sample methods aggregate signals across multiple generations, scoring uncertainty by their consistency, similarity, or variability. Representative examples include semantic \cite{farquhar2024detecting} and predictive \cite{kadavath2022language} entropy, conformal prediction \cite{kumar2023conformal}, and pairwise similarity methods \cite{lin2023generating}. Probing-based methods instead train lightweight predictors over internal activations to infer uncertainty directly from the model’s hidden representations \cite{azaria2023internal,dakhmouche2025can, liu2024uncertainty}.

\subsection{Selective Classification}

Selective classification \citep{chow1970optimum, selective_class_dnns} augments a predictor with the option to abstain, trading coverage for reduced error on the retained inputs. Formally, a selective classifier is a pair $(f, g)$, where $f : \mathcal{X} \to \mathcal{Y}$ is a predictor and $g : \mathcal{X} \to \{0, 1\}$ is a gating function: the prediction $f(x)$ is returned when $g(x) = 1$ and withheld otherwise. In practice, $g$ is induced by thresholding a confidence score $\kappa : \mathcal{X} \to \mathbb{R}$, so that performance reduces to the quality of $\kappa$ as a ranking of inputs by likely correctness.

Two quantities characterize such a classifier: \emph{coverage}, the fraction of inputs on which $f$ commits to a prediction, and \emph{selective risk}, the average loss on those inputs. Varying the threshold on $\kappa$ traces the risk-coverage curve~\citep{aurc}, whose area (AURC) summarizes performance across all operating points. Hence, selective classification provides a natural testbed for uncertainty estimators: a reliable confidence signal should report low selective risk across coverage levels \citep{ding2020revisiting}.

\section{Methodology}

\begin{figure}[t]
  \centering
  \includegraphics[width=\columnwidth]{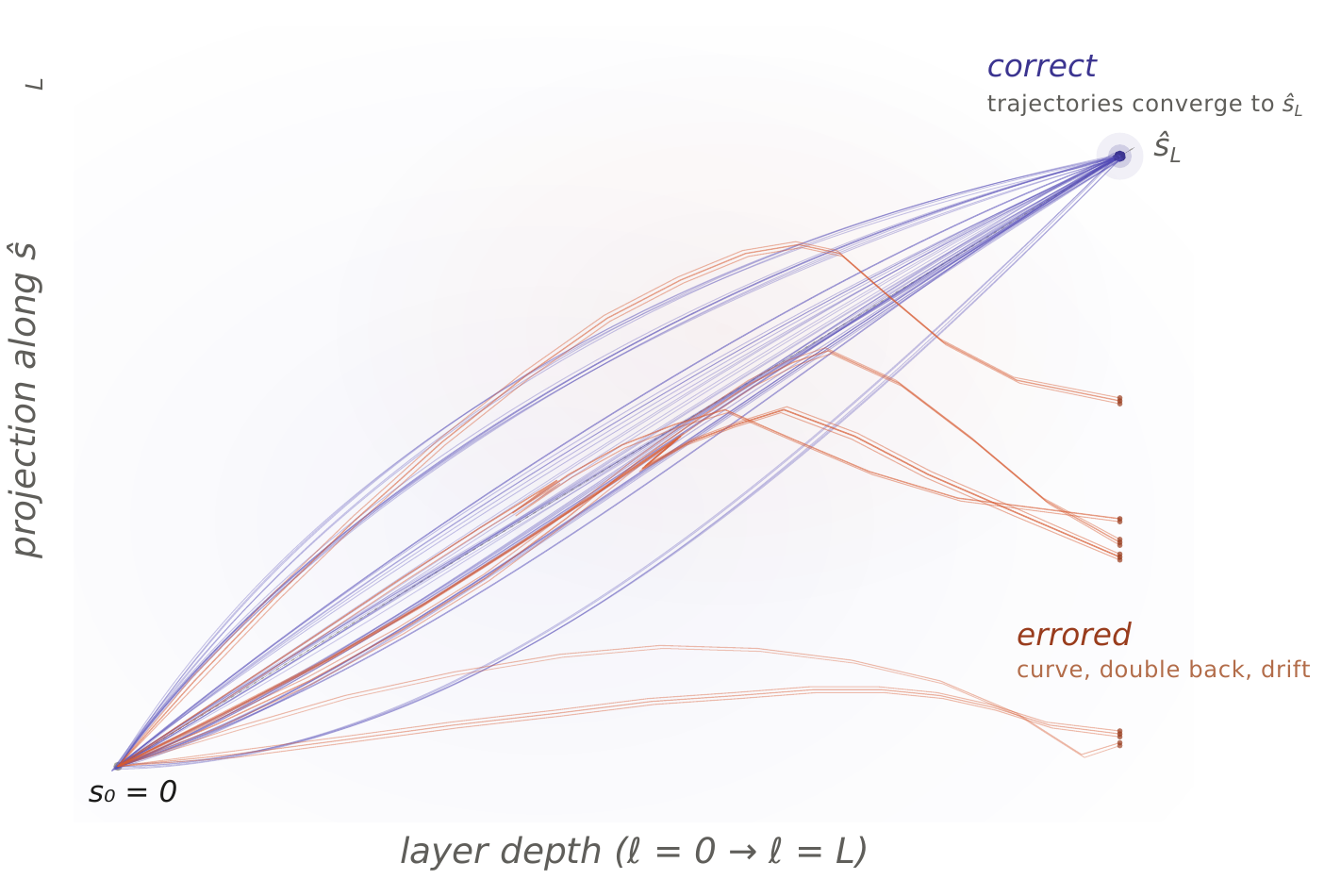}
  \caption{Cumulative MLP write-vectors traced layer-wise. The trajectory-geometry separates the two populations. Correct trajectories converge to $\hat{s}_L$; errored trajectories curve, double back, and drift.} 
  \label{fig:trajectories}
\end{figure}

We capture a language model's uncertainty by considering trajectory-information during sequential layer-wise processing. At each layer in a transformer architecture, the MLP contributes to the residual stream. The cumulative sum of these contributions forms a trajectory through representation space. We hypothesize that the geometry of these trajectories, including the distribution of update magnitudes across depth, local changes in direction, and the efficiency with which the path approaches its endpoint, contains information about uncertainty (Figure~\ref{fig:trajectories}). We summarize trajectory geometry using eleven scalar descriptors, which we combine with the model’s MSP in a sparse linear model.

The remainder of this section describes the representational substrate (\S\ref{sec:setup}), the extracted trajectory features (\S\ref{sec:features}), and the sparse linear probe (\S\ref{sec:probe}). Code to reproduce the experiments is available at \url{https://anonymous.4open.science/r/uq-motion-66CC/}. All experiments were conducted on a server equipped with an NVIDIA H100 NVL GPU.


\subsection{Setup and Substrate}
\label{sec:setup}


We consider finite-choice classification with a candidate set $\mathcal{Y}$, $|\mathcal{Y}|=K$. Given a prompt $x \in \mathcal{X}$, the model $f_\theta$ induces a distribution $p_\theta(y \mid x)$ over $\mathcal{Y}$ by applying a softmax to the next-token logits of the $K$ candidate-identifying tokens. From a single forward pass, we estimate an uncertainty score $u:\mathcal{X}\to[0,1]$ corresponding to the probability of error. We use the canonical MSP-based score $1-p_{\mathrm{msp}}(x)$, where $p_{\mathrm{msp}}(x)=\max_{y\in\mathcal{Y}} p_\theta(y\mid x)$, as the baseline \cite{hendrycks2016baseline}. We write $\hat{y}(x) = \arg\max_{y \in \mathcal{Y}} p_\theta(y \mid x)$ for the prediction and $e(x) = \mathbf{1}\bigl[\hat{y}(x) \neq y\bigr]$ for the error indicator, where $y$ denotes the ground-truth answer.



As the basis for $u$ we extract a layer-indexed sequence of MLP write-vectors at the final prompt position, the readout position used to compute the next-token logits over $\mathcal{Y}$. Each transformer block $\ell$ contains a multi-layer perceptron (MLP) sub-layer that writes a contribution $m_\ell(x) \in \mathbb{R}^H$ to the residual stream, where $H$ is the model's hidden dimension; we capture $m_\ell(x)$ for each block $\ell = 1, \dots, L$ via forward hooks placed on the MLP sub-module, recording its output at the final token of the prompt.



We focus on MLP write-vectors because prior mechanistic interpretability work identifies transformer MLPs as important sites for factual knowledge storage and recall \citep{geva2021transformer, meng2022locating, yu2024mechanistic}. This motivates using MLP write-vectors as the unit of analysis for studying how predictions are assembled across layers. We also consider the partial sums
\begin{align}
s_\ell(x) = \sum_{k \leq \ell} m_k(x),
\end{align}
where $m_k(x)$ is the layer-$k$ MLP write-vector. The sequence $\{s_\ell(x)\}_{\ell=1}^{L}$ traces a discrete trajectory in residual-stream space, with $s_L(x)$ the total MLP-driven displacement over the forward pass and $\hat{s}_L(x) = s_L(x) / \|s_L(x)\|$ its unit direction. We write $\mu = L^{-1} \sum_k \|m_k\|$ for the mean update norm and $T = \sum_k \|m_k\|$ for the total path length.

\subsection{Trajectory Features}
\label{sec:features}

We describe each trajectory $(m_\ell, s_\ell)_{\ell=1}^L$ via eleven layer-wise scalar features
(Table~\ref{tab:features}). The features are scale-invariant, enabling comparison across
models with different hidden dimensions, and fall into four geometric groups:

\begin{enumerate}[
  leftmargin=2.8em,
  labelsep=0.5em,
  itemsep=2pt,
  topsep=2pt,
  parsep=0pt,
  partopsep=0pt,
  label={}
]
  \item[\textcolor{gOneAcc}{(\textsc{G}1)}]   \feat{gOneAcc}{Depth allocation}: how
computational effort is distributed across layers;
  \item[\textcolor{gTwoAcc}{(\textsc{G}2)}]   \feat{gTwoAcc}{Local shape}: the local
geometry of the trajectory;
  \item[\textcolor{gThreeAcc}{(\textsc{G}3)}] \feat{gThreeAcc}{Endpoint alignment}: how each
layer relates to the trajectory's endpoint; and
  \item[\textcolor{gFourAcc}{(\textsc{G}4)}]  \feat{gFourAcc}{Trajectory efficiency}: how
directly the trajectory reaches its endpoint.
\end{enumerate}

\textcolor{gOneAcc}{G1} captures the depth distribution of update magnitude. The
\feat{gOneAcc}{relative update magnitude} flags layers whose contribution is
disproportionately large relative to the layer-wise average, while the
\feat{gOneAcc}{cumulative path fraction} reveals whether the model front- or back-loads its
computation.

\textcolor{gTwoAcc}{G2} characterizes local trajectory shape. The \feat{gTwoAcc}{consecutive
cosine} measures the directional consistency of successive updates, high values indicate a
smooth path through representation space, while the \feat{gTwoAcc}{curvature} highlights
abrupt changes of direction. The \feat{gTwoAcc}{update-state alignment} distinguishes layers
that reinforce prior computation from those that contradict it.

\textcolor{gThreeAcc}{G3} relates each layer to the trajectory's endpoint. Two cosines track
convergence: the \feat{gThreeAcc}{direction to final} measures whether the cumulative state
aligns with the endpoint, and the \feat{gThreeAcc}{update to final} measures whether
individual updates point toward it. The \feat{gThreeAcc}{signed final support} quantifies an update's alignment with the final direction. The negative part of the \emph{update to final}, the \feat{gThreeAcc}{contradictory support}, isolates layers that actively oppose the endpoint,
while the \feat{gThreeAcc}{orthogonal mass fraction} captures the share of an update's
magnitude orthogonal to the final direction.

Finally, \textcolor{gFourAcc}{G4} measures efficiency via the \feat{gFourAcc}{cumulative
coherence}, the ratio of net displacement to total path length: it equals one for a perfectly
straight trajectory and approaches zero when updates cancel rather than accumulate.

\subsection{Sparse Linear Probe}
\label{sec:probe}

We model the uncertainty score via sparse logistic regression:
\begin{align}
  u(x) &= \sigma\!\left(w^\top z(x) + b\right), \\
  z(x) &= \bigl[\, \varphi(x),\; p_{\mathrm{msp}}(x)\,\varphi(x) \,\bigr],
  \label{eq:probe}
\end{align}
where $\varphi(x) \in \mathbb{R}^{L \times 10}$ stacks the layer-wise features of
Section~\ref{sec:features} and $p_{\mathrm{msp}}(x)$ is the MSP of the model's prediction. The interaction term lets the probe weight each feature by confidence: curvature under a high-confidence prediction is a stronger error signal than the same curvature when the model is already uncertain. We minimize class-balanced binary cross-entropy against the error indicator $e(x)$ of Section~\ref{sec:setup} under an elastic-net penalty \cite{zou2005regularization}; sparsity in the fitted $w$ then identifies the predictive features directly, with no separate selection stage.

Each dataset is split 65/15/20 into train, validation, and test folds, stratified by the error
indicator so that the error rate is matched across folds. We sweep a 64-point grid over the
regularization strength $C$ and $\ell_1$ mixing ratio $\rho$
(Appendix~\ref{app:hyperparameters}), selecting the setting with lowest validation AURC. The
selected hyperparameters are refit on train$+$validation before evaluation on the held-out
test fold.

\section{Experimental Setup}
\label{sec:experimental-setup}

\begin{table*}[!t]
\centering
\small
\renewcommand{\arraystretch}{1.2}
\begin{tabular*}{\textwidth}{@{\extracolsep{\fill}}l
    S[table-format=2.2]@{\hskip 4pt}l
    S[table-format=2.2]@{\hskip 4pt}l
    S[table-format=2.2]@{\hskip 4pt}l
    S[table-format=2.2]@{\hskip 4pt}l
    S[table-format=2.2]@{\hskip 4pt}l@{}}
\toprule
& \multicolumn{2}{c}{\textbf{MMLU}}
& \multicolumn{2}{c}{\textbf{CosmosQA}}
& \multicolumn{2}{c}{\textbf{HellaSwag}}
& \multicolumn{2}{c}{\textbf{HaluDial}}
& \multicolumn{2}{c}{\textbf{HaluSum}}\\
\cmidrule(lr){2-3}\cmidrule(lr){4-5}\cmidrule(lr){6-7}\cmidrule(lr){8-9}\cmidrule(lr){10-11}
\textbf{Model} & {Ours} & {$\Delta$} & {Ours} & {$\Delta$} & {Ours} & {$\Delta$} & {Ours} & {$\Delta$} & {Ours} & {$\Delta$}\\
\midrule
Llama-3.2-3B   & 28.68 & \gain{21.83} &  6.83 & \gain{0.83}  & 31.99 & \gain{1.61}  & 38.17 & \gain{11.45} & 34.38 & \gain{16.54}\\
Llama-3.1-8B   & 22.65 & \gain{3.82}  &  3.58 & \gain{0.27}  & 26.87 & \gain{3.27}  & 22.48 & \gain{6.59}  & 21.13 & \gain{21.06}\\
Llama-3.3-70B  & 14.86 & \gain{8.74}  &  1.41 & \gain{0.37}  &  3.00 & \gain{1.71}  &  7.99 & \gain{5.41}  & 10.10 & \gain{20.17}\\
Qwen2.5-7B     & 14.87 & \gain{1.65}  &  3.94 & \loss{0.18}  &  6.59 & \gain{2.07}  & 19.74 & \gain{4.28}  & 20.67 & \gain{9.65}\\
Qwen2.5-14B    & 10.86 & \gain{0.23}  &  2.24 & \gain{0.16}  &  2.75 & \gain{0.38}  & 12.05 & \gain{3.94}  &  9.87 & \gain{16.25}\\
Qwen2-72B      &  6.10 & \gain{2.21}  &  1.27 & \gain{0.42}  &  1.75 & \gain{1.15}  &  7.29 & \gain{4.45}  & 10.84 & \gain{16.42}\\
Qwen2.5-72B    &  5.29 & \gain{1.96}  &  1.06 & \gain{0.79}  &  1.82 & \gain{2.75}  &  9.46 & \gain{7.50}  &  8.82 & \gain{16.74}\\
DeepSeek-7B    & 33.15 & \gain{1.55}  &  7.47 & \loss{0.16}  & 20.65 & \gain{5.32}  & 27.21 & \gain{6.58}  & 29.30 & \gain{14.29}\\
DeepSeek-67B   & 11.83 & \loss{0.20}  &  1.50 & \loss{0.01}  &  2.75 & \gain{0.28}  &  9.38 & \gain{5.35}  & 11.26 & \gain{20.18}\\
\bottomrule
\end{tabular*}
\vspace{4pt}
\caption{We report AURC ($\times 100$; lower is better) for our method, together with the absolute reduction $\Delta = \mathrm{MSP} - \mathrm{Ours}$ relative to the MSP baseline. Positive $\Delta$ ($\blacktriangle$) denotes improvement; negative $\Delta$ ($\blacktriangledown$) denotes regression. Full comparisons against the Ceiling and Ablation experiments are reported in Appendix~\ref{app:full-iid-aurc}.}
\label{tab:iid-aurc-main}
\vspace{-1em}
\end{table*}

\paragraph{Models.} We conduct the experiments using 9 instruction-tuned LLMs from three model families, spanning scales from 3B to 72B parameters: Qwen (Qwen2.5-7B-Instruct, Qwen2.5-14B-Instruct, Qwen2-72B-Instruct, Qwen2.5-72B-Instruct) \cite{Yang2024Qwen25TR},  Llama (Llama-3.2-3B-Instruct, Llama-3.1-8B-Instruct, Llama-3.3-70B-Instruct) \cite{grattafiori2024llama}, and DeepSeek (deepseek-llm-7b-chat, deepseek-llm-67b-chat) \cite{bi2024deepseek}.
\paragraph{Datasets.}  We use five benchmark datasets adapted from \citet{ye2024benchmarking} spanning question answering (MMLU; \citealp{hendrycks2020measuring}), reading comprehension (CosmosQA; \citealp{huang2019cosmos}), commonsense inference (HellaSwag; \citealp{zellers2019hellaswag}), dialogue response selection (HaluDial; \citealp{li2023halueval}), and document summarization (HaluSum; \citealp{li2023halueval}). Each dataset contains 10,000 instances formatted as four-option (A--D) multiple-choice questions in a zero-shot setting, using a base prompt that presents the question and options directly with the prefix \texttt{Answer:}. The predicted answer is the option with maximum softmax probability (MSP), which also serves as our primary baseline for uncertainty estimation.
\paragraph{Metrics.}  We evaluate the quality of our probe's uncertainty profile using  AURC ~\citep{aurc}. In essence, a reliable uncertainty profile should allow the model to abstain on observations it is likely to get wrong, so that errors decrease as we restrict predictions to the most confident observations.

\begin{figure*}[t]
    \centering
    
    \includegraphics[width=\textwidth]{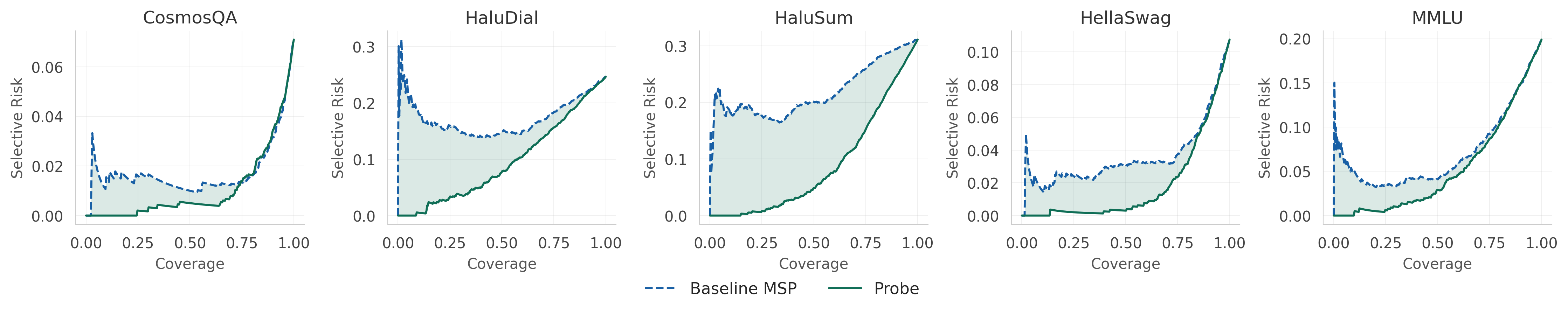}
    \caption{Risk-coverage curves for Qwen2.5-72B across the five evaluation datasets. An ideal curve is monotonically non-decreasing as predictions are rejected in order of decreasing confidence, approaching zero risk at low coverage and the base error rate at full coverage. The MSP baseline is shown in dashed blue and our probe in solid green; the shaded region between them indicates the probe's gain.}
    \label{fig:rc-curves-qwen}

    \vspace{1em}

    \includegraphics[width=\textwidth]{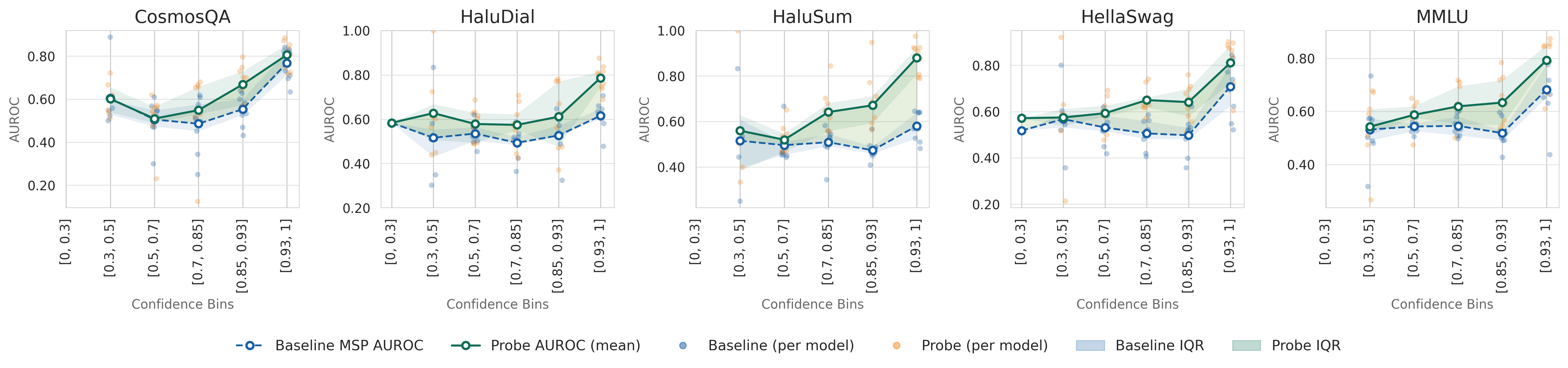}
    \caption{Probe and MSP AUROC across confidence bins, aggregated over models. Lines show mean AUROC, bands the interquartile range, and dots per-model values. Green shading indicates bins where the probe outperforms MSP.}
    \label{fig:auroc-across-conf-bins}
\end{figure*}

Let $f$ be a predictor with a confidence function $\kappa$, evaluated on $\{(x_i, y_i)\}_{i=1}^{n}$ under the $0/1$ loss, and re-index the samples so that $\kappa(x_1) \geq \dots \geq \kappa(x_n)$. The selective risk at coverage $c \in (0, 1]$ is the average loss on the top $\lceil c n \rceil$ samples, defined as
\begin{equation}
    R(c) \;=\; \frac{1}{\lceil c n \rceil} \sum_{i=1}^{\lceil c n \rceil} \mathbf{1}\bigl[f(x_i) \neq y_i\bigr],
\end{equation}
where AURC averages $R(c)$ over all coverage levels:
\begin{equation}
    \mathrm{AURC}(f, \kappa) \;=\; \frac{1}{n} \sum_{k=1}^{n} R\!\left(\tfrac{k}{n}\right).
\end{equation}
Lower AURC indicates that the confidence function more effectively separates correct from incorrect predictions, assigning higher confidence to the former. In confidence-stratified analyses (Figure~\ref{fig:auroc-across-conf-bins}) we additionally report AUROC, the area under the ROC curve for discriminating correct from incorrect predictions within each confidence bin.

\paragraph{Relation to calibration.}

Expected Calibration Error (ECE)~\cite{naeini2015obtaining} assesses whether confidence values match empirical accuracy, but calibration and selective performance are distinct~\cite{ding2020revisiting}. A constant score equal to the empirical accuracy can achieve ECE $=0$ while providing no useful ranking for abstention. Conversely, a score can induce a near-optimal ranking for selective prediction while being numerically miscalibrated. Since AURC directly evaluates the risk-coverage trade-off determined by the ordering induced by $\kappa$, we use it as our primary metric.

\paragraph{Baselines.}

We compare our probe with two baselines and one ablation. The first baseline is maximum softmax probability (MSP), the probability $p_{\mathrm{msp}}(x)$ assigned to the predicted token, which we convert to the uncertainty score $1 - p_{\mathrm{msp}}(x)$. MSP requires no auxiliary computation beyond the model forward pass and is a strong standard baseline for confidence estimation~\citep{vashurin2025benchmarking, dakhmouche2025can}. The second baseline is a raw-activation probe in the spirit of \citet{azaria2023internal}, extended to all layers: we replace our eleven trajectory features with the per-layer hidden states and train a layer-weighted MLP on binary error labels. This provides a high-capacity reference for the error signal recoverable from activations, without the interpretability constraints of our feature-based probe. We refer to this baseline as the \emph{ceiling}. The \emph{trajectory-only} ablation restricts the input to $\varphi(x)$, removing MSP and its interaction terms; its gap to the full probe quantifies the complementarity between trajectory geometry and MSP.

\section{Results}

We compare the probe with the baselines and ablation from Section~\ref{sec:experimental-setup}, reporting headline results in Table~\ref{tab:iid-aurc-main} and full results in Appendix Table~\ref{tab:iid-aurc-full}. We then assess its uncertainty profile and the geometric signals it uses across depth.


\paragraph{Probe performance.}
Our probe improves over MSP on 41 of 45 model-dataset pairs, with gains largest where MSP performs worst (Spearman $r_s=0.78$). For the five configurations with MSP AURC above 40, it reduces AURC by 11.45--21.83 points, with the largest reductions on Llama-3.2-3B MMLU, Llama-3.1-8B HaluSum, and Llama-3.2-3B HaluSum. The probe also matches or outperforms the ceiling on 24 configurations and is within 2 AURC points on another 6, despite using only eleven scalar features instead of full hidden states. Finally, the \emph{trajectory-only} ablation outperforms MSP on 35 configurations, showing that trajectory geometry is independently informative, while the full probe further benefits from MSP as a complementary signal.

Figure~\ref{fig:rc-curves-qwen} dissects these AURC gains into risk-coverage curves for Qwen2.5-72B. The probe (solid green) lies at or below MSP (dashed blue) throughout the coverage range, with the largest gap on HaluDial and HaluSum. MSP exhibits an unstable low-coverage spike on every dataset, whereas the probe rises smoothly from zero and sustains a low-risk regime up to roughly 50--60\% coverage.

Figure~\ref{fig:auroc-across-conf-bins} stratifies AUROC by MSP confidence. Across all five datasets, MSP performs close to random-chance levels (AUROC $\approx 0.5$--$0.6$) outside its highest-confidence bin. In contrast, the probe consistently outperforms MSP across nearly all bins, with the largest improvements emerging in the mid and high-confidence regimes for HaluSum, HaluDial, HellaSwag, and MMLU.

\begin{figure}[H]
    \centering
    \includegraphics[width=\linewidth]{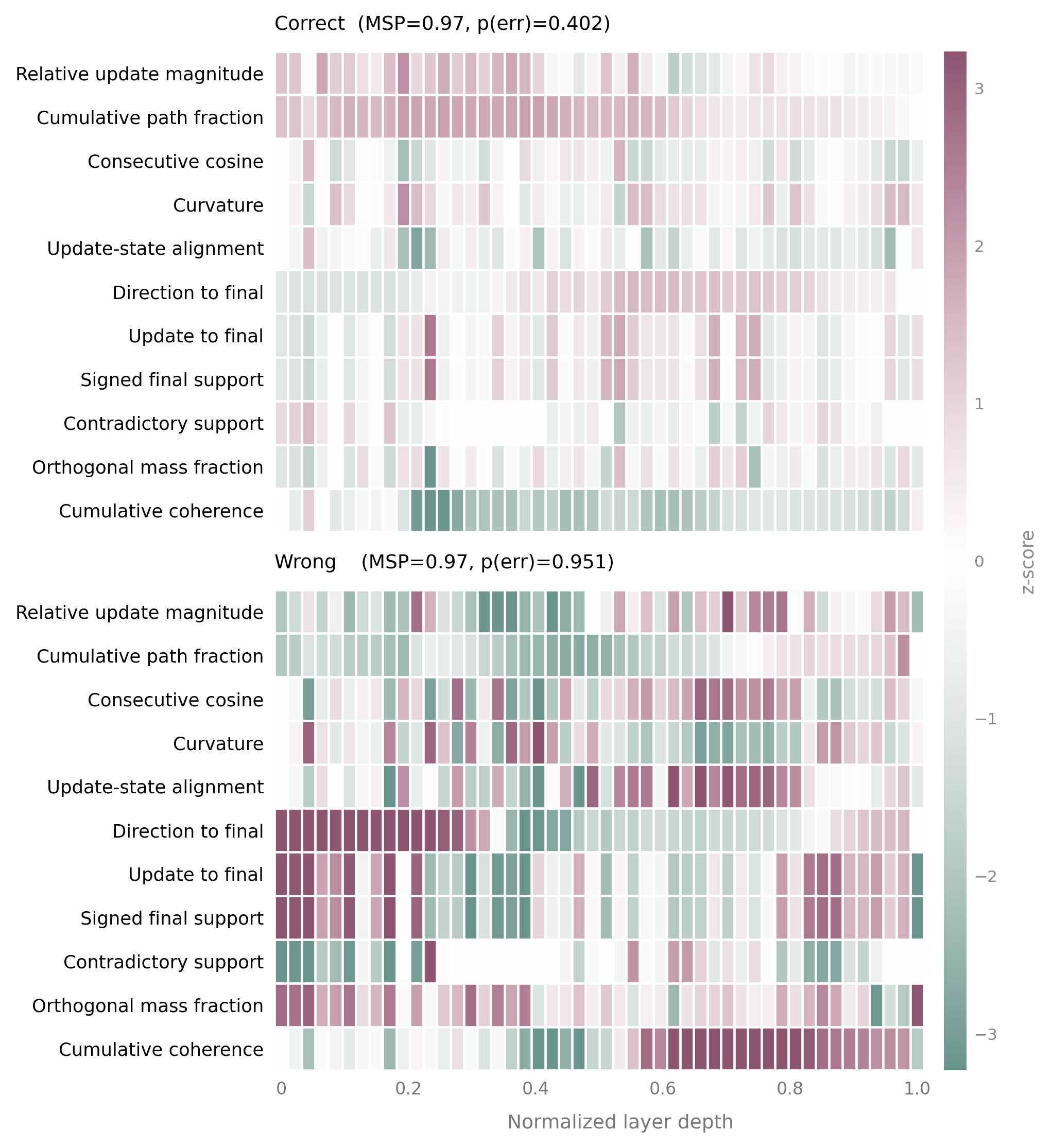}
    \caption{Two Qwen2.5-14B predictions on MMLU at MSP $\approx 0.97$, one correct and one incorrect. Layer-wise z-scores of the eleven trajectory features across normalized depth for the correct prediction (top) and incorrect (bottom).}
    \label{fig:geometry_pairs}
\end{figure}

To examine whether the probe finds geometric signal at fixed MSP, Figure~\ref{fig:geometry_pairs} compares two Qwen2.5-14B predictions on MMLU with MSP $\approx 0.97$ -- one correct, one incorrect -- via the eleven trajectory features z-scored against the per-layer population. In early layers (depth $0$--$0.3$), the incorrect prediction has already aligned with its final direction (\emph{Direction to final} and \emph{Update to final} saturated at $z \approx +3$), while the correct prediction shows none of this early alignment. The correct prediction also front-loads its computation (\emph{Cumulative path fraction} above the population mean at $0$--$0.5$), whereas the incorrect prediction back-loads its path length and traces an unusually straight cumulative path in the second half (\emph{Cumulative coherence} positive at depths $0.55$--$0.85$).

\begin{figure}[H]
    \centering
    \includegraphics[width=\linewidth, trim=0 0 0 3, clip]{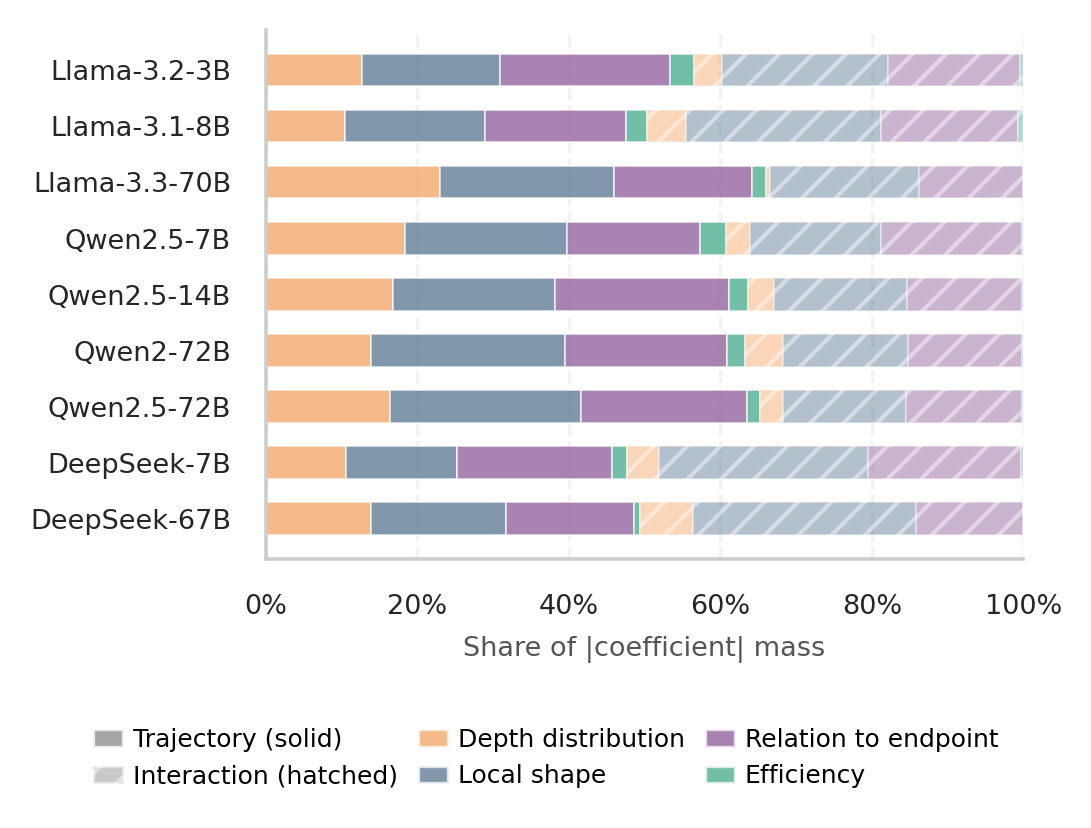}
    \caption{Probe coefficient composition across models, showing coefficient mass by geometric feature family (color), split into main effects (solid) and MSP$\times$trajectory interactions (hatched).}
    \label{fig:family_composition}
\end{figure}

\paragraph{Probe interpretability.}
We examine what signal the probe uses by decomposing its total coefficient mass across the four feature groups in Table~\ref{tab:features}, separating mass assigned to raw trajectory features $\varphi(x)$ from that assigned to their MSP interactions, $p_{\mathrm{msp}} \cdot \varphi(x)$. Figure~\ref{fig:family_composition} reports this decomposition by model, aggregated across datasets since the pattern is stable; solid bars indicate trajectory features and hatched bars indicate interactions. The two blocks contribute roughly equally overall. Trajectory features carry a larger share for Llama-3.3-70B, Qwen2.5-14B, Qwen2-72B, and Qwen2.5-72B, whereas DeepSeek-7B and Llama-3.1-8B assign relatively less mass to depth-distribution features. Depth distribution is also the only feature group whose mass lies mainly on the raw features rather than on MSP interactions; the remaining groups split their mass more evenly.

\begin{figure}[H]
    \centering
    \includegraphics[width=\linewidth]{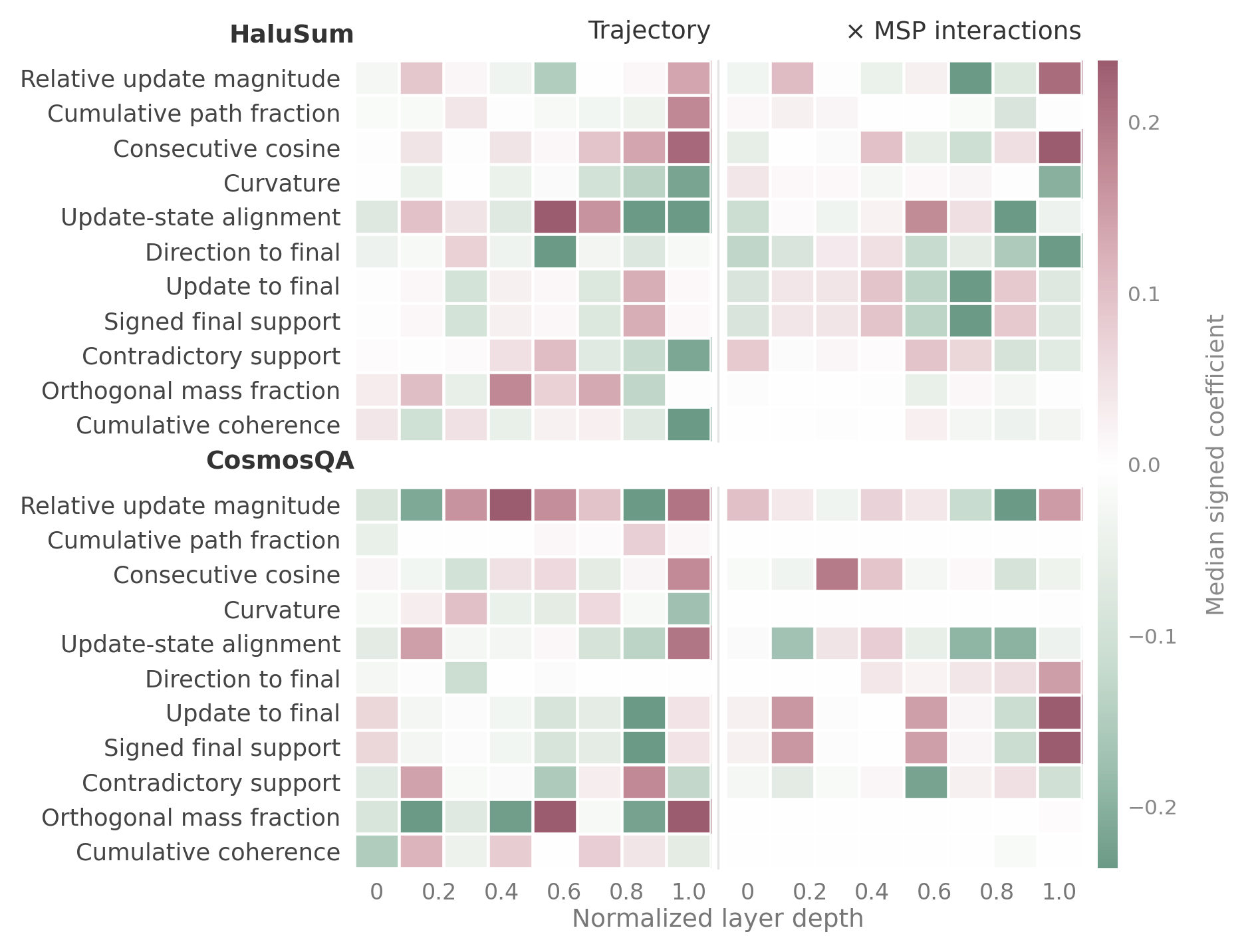}
    \caption{Median probe coefficients per (feature, depth-bin) cell for Qwen2.5-14B on HaluSum (top) and CosmosQA (bottom). Left: direct effects; right: MSP interactions. Pink increases predicted error probability, green decreases it. Values are normalized by the dataset-specific 95th-percentile magnitude.}
\label{fig:depth_profile}
\end{figure}

Figure~\ref{fig:depth_profile} maps the median probe coefficient at each (feature, depth) cell for Qwen2.5-14B on HaluSum and CosmosQA, with direct effects $\varphi(x)$ on the left and MSP interactions on the right. The two datasets show distinct signatures. On HaluSum, signal concentrates in the final quarter (depth $\gtrsim 0.75$): errors carry a large, smooth late update-positive weight on \emph{Relative update magnitude}, \emph{Consecutive cosine}, and \emph{Cumulative path fraction} near the output -- counterbalanced by negative \emph{Update-state alignment} and \emph{Cumulative coherence}, indicating that updates aligned with the running state and efficient cumulative paths predict correctness rather than error. \emph{Direction to final} is also negative at depth $0.5$--$0.6$, so trajectories that align with their endpoint earlier are less error-prone. HaluSum errors reflect premature commitment followed by a late, state-breaking correction. CosmosQA shows no such localization: coefficients are smaller and spread across depth, the strongest cell at mid-depth (\emph{Orthogonal mass  fraction} at $0.5$--$0.6$), so error trajectories drift sideways relative to the endpoint. Endpoint-alignment features (\emph{Update to final}, \emph{Signed final support}, \emph{Direction to final}) light up in the MSP-interaction panel near the output, predicting error only when weighted by confidence.

\begin{figure}[H]
    \centering
    \includegraphics[width=\linewidth]{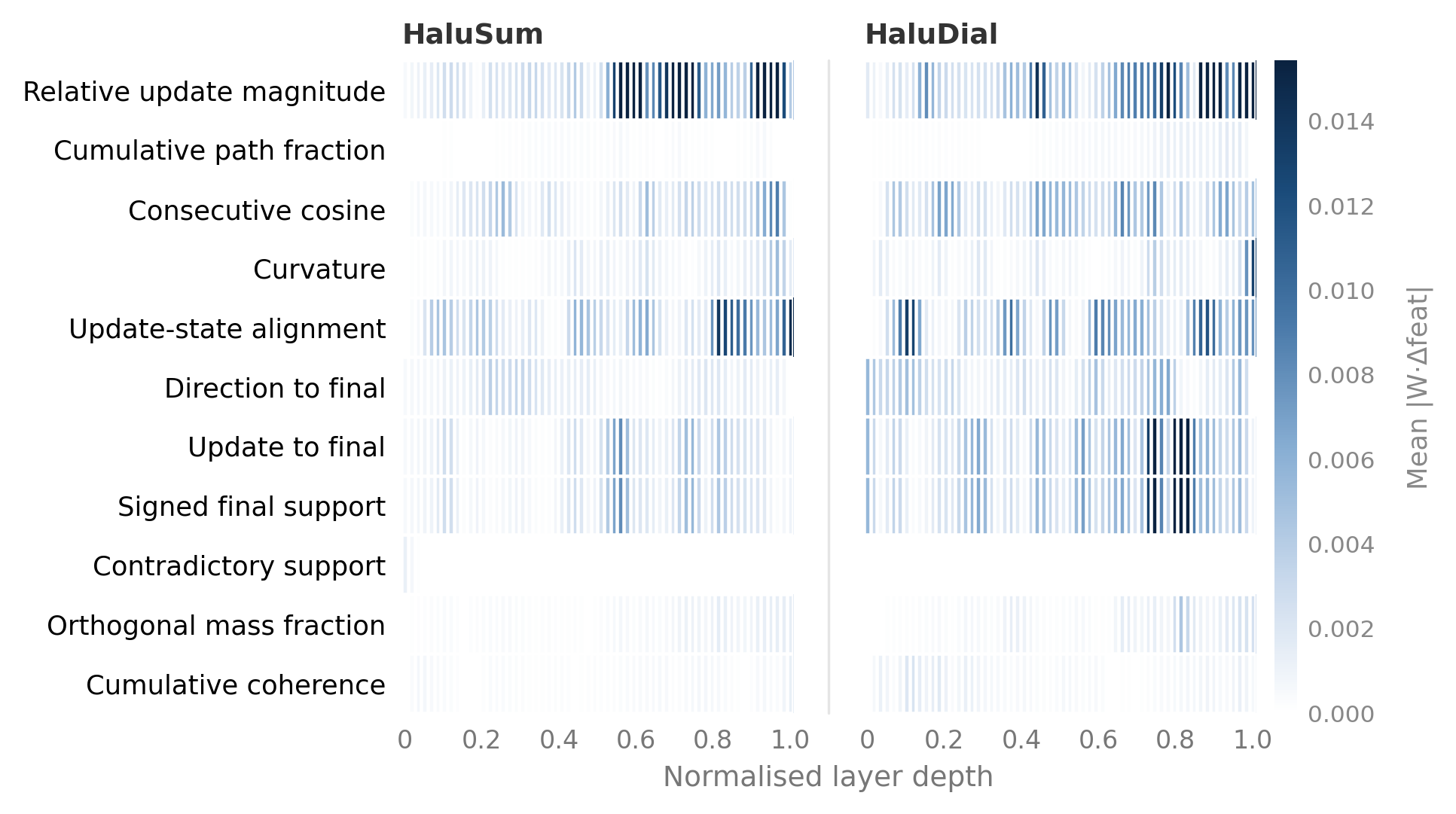}
    \caption{Aggregate attribution maps for Llama-3.2-3B-Instruct on HaluSum (left) and HaluDial (right), averaged over the 100 MSP-matched probe-correct pairs with the largest error-probability gaps. Maps share a normalized-depth axis and color scale.}
\label{fig:attribution_map}
\end{figure}

The coefficient maps in Figure~\ref{fig:depth_profile} describe the probe's learned structure; to see it deployed, we turn to the pairs on which MSP confidence is silent (Figure \ref{fig:attribution_map}). Within the top 30\% most-confident Llama-3.2-3B-Instruct predictions, we match each probe-flagged error to a probe-cleared non-error with MSP agreeing within 0.02. For each (feature, depth) cell, we multiply the feature-value difference between the paired examples by the probe's effective coefficient at the shared confidence (so both trajectory and MSP-interaction blocks contribute), then average absolute values across pairs. A cell is bright only when trajectories diverge there \emph{and} the probe weights it, localizing signal MSP alone misses. On both datasets the signal concentrates in the second half of the network, with \emph{Relative update magnitude} dominant -- engaging from depth $\sim\!0.5$ on HaluSum but only near depth $1.0$ on HaluDial. HaluSum additionally leans on \emph{Update-state alignment} near the output, while HaluDial shifts to the endpoint-alignment channel (\emph{Update to final}, \emph{Signed final support}) at depth $0.7$--$0.85$.

\section{Discussion}
We find that the trajectory of a model's layer-wise computation reveals when its answer should be trusted. Correct predictions and confident errors can be indistinguishable in the output distribution yet diverge in the geometry that produced them, and a few interpretable
descriptors of that geometry suffice to recover the distinction. A sparse linear probe trained on them improves over MSP across nearly all configurations, with gains largest where MSP is most miscalibrated; on more than half of those configurations, the eleven scalar features match a probe trained on the the full hidden states. The resulting selective prediction function is well-behaved: the probe traces a smooth, monotone risk-coverage curve, sustaining low risk where MSP tends to spike at low coverage.
Because the features are interpretable by construction, the probe's coefficients localize \emph{how} a model fails, not just \emph{whether} it does. On HaluSum, errors take the form of premature endpoint commitment in the first half of the network, followed by a late, oversized MLP write that breaks the running state. On CosmosQA, errors instead manifest as mid-depth drift orthogonal to the eventual answer direction. These mechanisms remain legible even when MSP fails to discriminate: among predictions matched in confidence, correct and erroneous trajectories diverge at specific (feature, depth) cells the probe weights, localizing discriminative structure that the output distribution has flattened away. The signatures, however, are task-specific, in line with the recent finding that “geometries of truth” are largely orthogonal across tasks \cite{azizian2025geometries}. Together, these results point to a practical path to trustworthy selective prediction: the features add negligible cost at inference, and their interpretability makes failures auditable rather than opaque. \textbf{Limitations.} Our analysis is restricted to discrete-choice settings, where the prediction is localized to a single answer position. The probe is also fit per (model, task), and preliminary leave-one-dataset-out experiments show gains over MSP on only a subset of configurations. Identifying trajectory features that transfer across models and tasks remains open.



\section{Related Work}

Probing assesses what information is linearly encoded in neural network representations \cite{alain2016understanding} and language models \cite{petroni2019language, azaria2023internal}: a classifier is trained on activations extracted from a frozen model to predict a target property. \citet{azaria2023internal} introduced this approach for UQ in LLMs, showing that internal representations contain reliable signals of output correctness. Subsequent work on the ``geometry of truth'' suggests a direction in model internals along which truthful and erroneous outputs are approximately linearly separable \cite{azizian2025geometries, li2023inference, marks2023geometry, kossen2024semantic, dakhmouche2025can, liu2024uncertainty, beigi2024internalinspector, burns2022discovering}, differing mainly in how this direction is recovered. \citet{li2023inference} train linear probes on individual attention heads for detection and inference-time steering. \citet{burns2022discovering} propose Contrast Consistent Search (CCS), recovering the direction without labels by enforcing consistency between a statement and its negation. 
\citet{dakhmouche2025can} take a Bayesian route, fitting layer-to-layer linear maps and combining their posterior log-likelihoods via sparse regression. In contrast, we extract trajectory geometry rather than classifying individual hidden states: eleven interpretable scalar descriptors of how MLP contributions accumulate across depth, fed to a sparse linear probe whose inputs and decision rule are interpretable end to end.

\section{Conclusion}
We introduced a method for reading uncertainty from the motion of language model computation. By tracing the answer-position residual stream as a cumulative path of per-layer MLP write-vectors and condensing it through eleven scale-invariant geometric descriptors, we obtain an uncertainty signal whose inputs and decision rule are inspectable end to end. Each feature has a direct geometric reading, computed in closed form from the activations with no auxiliary model, and the sparse probe trained on them localizes not only \emph{whether} a model is likely to err but \emph{how} -- which layers commit prematurely, which contradict the running state, where trajectories drift away from their endpoint. Treating activations as motion rather than as snapshots offers a tractable middle path between the opacity of dense probes and the single scalar of MSP, recovering signal from how a prediction is assembled, not only from where it lands. We hope this work encourages the development of simpler, more interpretable representations for uncertainty quantification in LLMs.




\section*{Impact Statement}

UQ is a core component of trustworthy AI, enabling LLMs to flag predictions that should be deferred, reviewed, or abstained from in high-stakes settings. Our method reads uncertainty from the geometry of MLP write-vectors to the residual stream within a single forward pass, offering an efficient and interpretable alternative to costly sampling-based or ensemble approaches. Because each feature has a closed-form geometric meaning, the resulting confidence signal is auditable end-to-end, allowing practitioners to inspect not only whether a model is uncertain but where in its computation that uncertainty arises.


\bibliography{example_paper}

@article{Yang2024Qwen25TR,
  title={Qwen2.5 Technical Report},
  author={Qwen An Yang and Baosong Yang and Beichen Zhang and Binyuan Hui and Bo Zheng and Bowen Yu and Chengyuan Li and Dayiheng Liu and Fei Huang and Guanting Dong and Haoran Wei and Huan Lin and Jian Yang and Jianhong Tu and Jianwei Zhang and Jianxin Yang and Jiaxin Yang and Jingren Zhou and Junyang Lin and Kai Dang and Keming Lu and Keqin Bao and Kexin Yang and Le Yu and Mei Li and Mingfeng Xue and Pei Zhang and Qin Zhu and Rui Men and Runji Lin and Tianhao Li and Tingyu Xia and Xingzhang Ren and Xuancheng Ren and Yang Fan and Yang Su and Yi-Chao Zhang and Yunyang Wan and Yuqi Liu and Zeyu Cui and Zhenru Zhang and Zihan Qiu and Shanghaoran Quan and Zekun Wang},
  journal={ArXiv},
  year={2024},
  volume={abs/2412.15115},
  url={https://api.semanticscholar.org/CorpusID:274859421}
}

@inproceedings{on_the_calibration_of_modern_neural_networks,
author = {Guo, Chuan and Pleiss, Geoff and Sun, Yu and Weinberger, Kilian Q.},
title = {On calibration of modern neural networks},
year = {2017},
publisher = {JMLR.org},
booktitle = {Proceedings of the 34th International Conference on Machine Learning - Volume 70},
pages = {1321–1330},
numpages = {10},
location = {Sydney, NSW, Australia},
series = {ICML'17}
}

@article{grattafiori2024llama,
  title={The llama 3 herd of models},
  author={Grattafiori, Aaron and Dubey, Abhimanyu and Jauhri, Abhinav and Pandey, Abhinav and Kadian, Abhishek and Al-Dahle, Ahmad and Letman, Aiesha and Mathur, Akhil and Schelten, Alan and Vaughan, Alex and others},
  journal={arXiv preprint arXiv:2407.21783},
  year={2024}
}

@article{bi2024deepseek,
  title={Deepseek llm: Scaling open-source language models with longtermism},
  author={Bi, Xiao and Chen, Deli and Chen, Guanting and Chen, Shanhuang and Dai, Damai and Deng, Chengqi and Ding, Honghui and Dong, Kai and Du, Qiushi and Fu, Zhe and others},
  journal={arXiv preprint arXiv:2401.02954},
  year={2024}
}

@article{ye2024benchmarking,
  title={Benchmarking llms via uncertainty quantification},
  author={Ye, Fanghua and Yang, Mingming and Pang, Jianhui and Wang, Longyue and Wong, Derek F and Yilmaz, Emine and Shi, Shuming and Tu, Zhaopeng},
  journal={Advances in Neural Information Processing Systems},
  volume={37},
  pages={15356--15385},
  year={2024}
}

@article{hendrycks2020measuring,
  title={Measuring massive multitask language understanding},
  author={Hendrycks, Dan and Burns, Collin and Basart, Steven and Zou, Andy and Mazeika, Mantas and Song, Dawn and Steinhardt, Jacob},
  journal={arXiv preprint arXiv:2009.03300},
  year={2020}
}

@inproceedings{manakul2023selfcheckgpt,
  title={Selfcheckgpt: Zero-resource black-box hallucination detection for generative large language models},
  author={Manakul, Potsawee and Liusie, Adian and Gales, Mark},
  booktitle={Proceedings of the 2023 conference on empirical methods in natural language processing},
  pages={9004--9017},
  year={2023}
}

@inproceedings{naeini2015obtaining,
  title={Obtaining well calibrated probabilities using bayesian binning},
  author={Naeini, Mahdi Pakdaman and Cooper, Gregory and Hauskrecht, Milos},
  booktitle={Proceedings of the AAAI conference on artificial intelligence},
  volume={29},
  number={1},
  year={2015}
}

@article{lin2023generating,
  title={Generating with confidence: Uncertainty quantification for black-box large language models},
  author={Lin, Zhen and Trivedi, Shubhendu and Sun, Jimeng},
  journal={arXiv preprint arXiv:2305.19187},
  year={2023}
}

@article{kumar2023conformal,
  title={Conformal prediction with large language models for multi-choice question answering},
  author={Kumar, Bhawesh and Lu, Charlie and Gupta, Gauri and Palepu, Anil and Bellamy, David and Raskar, Ramesh and Beam, Andrew},
  journal={arXiv preprint arXiv:2305.18404},
  year={2023}
}

@inproceedings{liu2025uncertainty,
  title={Uncertainty quantification and confidence calibration in large language models: A survey},
  author={Liu, Xiaoou and Chen, Tiejin and Da, Longchao and Chen, Chacha and Lin, Zhen and Wei, Hua},
  booktitle={Proceedings of the 31st ACM SIGKDD Conference on Knowledge Discovery and Data Mining V. 2},
  pages={6107--6117},
  year={2025}
}

@article{kuhn2023semantic,
  title={Semantic uncertainty: Linguistic invariances for uncertainty estimation in natural language generation},
  author={Kuhn, Lorenz and Gal, Yarin and Farquhar, Sebastian},
  journal={arXiv preprint arXiv:2302.09664},
  year={2023}
}

@article{vashurin2025benchmarking,
  title={Benchmarking uncertainty quantification methods for large language models with lm-polygraph},
  author={Vashurin, Roman and Fadeeva, Ekaterina and Vazhentsev, Artem and Rvanova, Lyudmila and Vasilev, Daniil and Tsvigun, Akim and Petrakov, Sergey and Xing, Rui and Sadallah, Abdelrahman and Grishchenkov, Kirill and others},
  journal={Transactions of the Association for Computational Linguistics},
  volume={13},
  pages={220--248},
  year={2025}
}

@inproceedings{petroni2019language,
  title={Language models as knowledge bases?},
  author={Petroni, Fabio and Rockt{\"a}schel, Tim and Riedel, Sebastian and Lewis, Patrick and Bakhtin, Anton and Wu, Yuxiang and Miller, Alexander},
  booktitle={Proceedings of the 2019 conference on empirical methods in natural language processing and the 9th international joint conference on natural language processing (EMNLP-IJCNLP)},
  pages={2463--2473},
  year={2019}
}

@article{alain2016understanding,
  title={Understanding intermediate layers using linear classifier probes},
  author={Alain, Guillaume and Bengio, Yoshua},
  journal={arXiv preprint arXiv:1610.01644},
  year={2016}
}

@inproceedings{beigi2024internalinspector,
  title={Internalinspector i2: Robust confidence estimation in llms through internal states},
  author={Beigi, Mohammad and Shen, Ying and Yang, Runing and Lin, Zihao and Wang, Qifan and Mohan, Ankith and He, Jianfeng and Jin, Ming and Lu, Chang-Tien and Huang, Lifu},
  booktitle={Findings of the association for computational linguistics: EMNLP 2024},
  pages={12847--12865},
  year={2024}
}

@article{kossen2024semantic,
  title={Semantic entropy probes: Robust and cheap hallucination detection in llms},
  author={Kossen, Jannik and Han, Jiatong and Razzak, Muhammed and Schut, Lisa and Malik, Shreshth and Gal, Yarin},
  journal={arXiv preprint arXiv:2406.15927},
  year={2024}
}

@article{marks2023geometry,
  title={The geometry of truth: Emergent linear structure in large language model representations of true/false datasets},
  author={Marks, Samuel and Tegmark, Max},
  journal={arXiv preprint arXiv:2310.06824},
  year={2023}
}

@article{azizian2025geometries,
  title={The Geometries of Truth Are Orthogonal Across Tasks},
  author={Azizian, Waiss and Kirchhof, Michael and Ndiaye, Eugene and Bethune, Louis and Klein, Michal and Ablin, Pierre and Cuturi, Marco},
  journal={arXiv preprint arXiv:2506.08572},
  year={2025}
}

@article{li2023inference,
  title={Inference-time intervention: Eliciting truthful answers from a language model},
  author={Li, Kenneth and Patel, Oam and Vi{\'e}gas, Fernanda and Pfister, Hanspeter and Wattenberg, Martin},
  journal={Advances in Neural Information Processing Systems},
  volume={36},
  pages={41451--41530},
  year={2023}
}

@inproceedings{margatina2023active,
  title={Active learning principles for in-context learning with large language models},
  author={Margatina, Katerina and Schick, Timo and Aletras, Nikolaos and Dwivedi-Yu, Jane},
  booktitle={Findings of the Association for Computational Linguistics: EMNLP 2023},
  pages={5011--5034},
  year={2023}
}

@article{hendrycks2016baseline,
  title={A baseline for detecting misclassified and out-of-distribution examples in neural networks},
  author={Hendrycks, Dan and Gimpel, Kevin},
  journal={arXiv preprint arXiv:1610.02136},
  year={2016}
}

@inproceedings{geva2021transformer,
  title={Transformer feed-forward layers are key-value memories},
  author={Geva, Mor and Schuster, Roei and Berant, Jonathan and Levy, Omer},
  booktitle={Proceedings of the 2021 Conference on Empirical Methods in Natural Language Processing},
  pages={5484--5495},
  year={2021}
}

@article{meng2022locating,
  title={Locating and editing factual associations in gpt},
  author={Meng, Kevin and Bau, David and Andonian, Alex and Belinkov, Yonatan},
  journal={Advances in neural information processing systems},
  volume={35},
  pages={17359--17372},
  year={2022}
}

@inproceedings{yu2024mechanistic,
  title={Mechanistic understanding and mitigation of language model non-factual hallucinations},
  author={Yu, Lei and Cao, Meng and Cheung, Jackie CK and Dong, Yue},
  booktitle={Findings of the Association for Computational Linguistics: EMNLP 2024},
  pages={7943--7956},
  year={2024}
}

@inproceedings{li2026semantic,
  title={Semantic volume: Quantifying and detecting both external and internal uncertainty in llms},
  author={Li, Xiaomin and Yu, Zhou and Zhang, Ziji and Zhuang, Yingying and Shah, Swair and Sadagopan, Narayanan and Beniwal, Anurag},
  booktitle={Proceedings of the AAAI Conference on Artificial Intelligence},
  volume={40},
  number={37},
  pages={31751--31759},
  year={2026}
}

@article{burns2022discovering,
  title={Discovering latent knowledge in language models without supervision},
  author={Burns, Collin and Ye, Haotian and Klein, Dan and Steinhardt, Jacob},
  journal={arXiv preprint arXiv:2212.03827},
  year={2022}
}

@inproceedings{ding2020revisiting,
  title={Revisiting the evaluation of uncertainty estimation and its application to explore model complexity-uncertainty trade-off},
  author={Ding, Yukun and Liu, Jinglan and Xiong, Jinjun and Shi, Yiyu},
  booktitle={Proceedings of the IEEE/CVF Conference on Computer Vision and Pattern Recognition Workshops},
  pages={4--5},
  year={2020}
}

@article{zou2005regularization,
  title={Regularization and variable selection via the elastic net},
  author={Zou, Hui and Hastie, Trevor},
  journal={Journal of the Royal Statistical Society Series B: Statistical Methodology},
  volume={67},
  number={2},
  pages={301--320},
  year={2005},
  publisher={Oxford University Press}
}

@inproceedings{
aurc,
title={Bias-Reduced Uncertainty Estimation for Deep Neural Classifiers},
author={Yonatan Geifman and Guy Uziel and Ran El-Yaniv},
booktitle={International Conference on Learning Representations},
year={2019},
url={https://openreview.net/forum?id=SJfb5jCqKm},
}

@inproceedings{azaria2023internal,
  title={The internal state of an LLM knows when it’s lying},
  author={Azaria, Amos and Mitchell, Tom},
  booktitle={Findings of the Association for Computational Linguistics: EMNLP 2023},
  pages={967--976},
  year={2023}
}

@article{liu2024uncertainty,
  title={Uncertainty estimation and quantification for llms: A simple supervised approach},
  author={Liu, Linyu and Pan, Yu and Li, Xiaocheng and Chen, Guanting},
  journal={arXiv preprint arXiv:2404.15993},
  year={2024}
}

@article{kendall2017uncertainties,
  title={What uncertainties do we need in bayesian deep learning for computer vision?},
  author={Kendall, Alex and Gal, Yarin},
  journal={Advances in neural information processing systems},
  volume={30},
  year={2017}
}

@article{kadavath2022language,
  title={Language models (mostly) know what they know},
  author={Kadavath, Saurav and Conerly, Tom and Askell, Amanda and Henighan, Tom and Drain, Dawn and Perez, Ethan and Schiefer, Nicholas and Hatfield-Dodds, Zac and DasSarma, Nova and Tran-Johnson, Eli and others},
  journal={arXiv preprint arXiv:2207.05221},
  year={2022}
}

@article{farquhar2024detecting,
  title={Detecting hallucinations in large language models using semantic entropy},
  author={Farquhar, Sebastian and Kossen, Jannik and Kuhn, Lorenz and Gal, Yarin},
  journal={Nature},
  volume={630},
  number={8017},
  pages={625--630},
  year={2024},
  publisher={Nature Publishing Group UK London}
}

@article{dakhmouche2025can,
  title={Can Linear Probes Measure LLM Uncertainty?},
  author={Dakhmouche, Ramzi and Letellier, Adrien and Gorji, Hossein},
  journal={arXiv preprint arXiv:2510.04108},
  year={2025}
}

@inproceedings{selective_class_dnns,
author = {Geifman, Yonatan and El-Yaniv, Ran},
title = {Selective classification for deep neural networks},
year = {2017},
isbn = {9781510860964},
publisher = {Curran Associates Inc.},
address = {Red Hook, NY, USA},
booktitle = {Proceedings of the 31st International Conference on Neural Information Processing Systems},
pages = {4885–4894},
numpages = {10},
location = {Long Beach, California, USA},
series = {NIPS'17}
}

@article{chow1970optimum,
  title={On optimum recognition error and reject tradeoff},
  author={Chow, C},
  journal={IEEE Transactions on information theory},
  volume={16},
  number={1},
  pages={41--46},
  year={1970},
  publisher={IEEE}
}

@inproceedings{huang2019cosmos,
  title={Cosmos QA: Machine reading comprehension with contextual commonsense reasoning},
  author={Huang, Lifu and Le Bras, Ronan and Bhagavatula, Chandra and Choi, Yejin},
  booktitle={Proceedings of the 2019 conference on empirical methods in natural language processing and the 9th international joint conference on natural language processing (EMNLP-IJCNLP)},
  pages={2391--2401},
  year={2019}
}

@inproceedings{zellers2019hellaswag,
  title={Hellaswag: Can a machine really finish your sentence?},
  author={Zellers, Rowan and Holtzman, Ari and Bisk, Yonatan and Farhadi, Ali and Choi, Yejin},
  booktitle={Proceedings of the 57th annual meeting of the association for computational linguistics},
  pages={4791--4800},
  year={2019}
}

@inproceedings{li2023halueval,
  title={Halueval: A large-scale hallucination evaluation benchmark for large language models},
  author={Li, Junyi and Cheng, Xiaoxue and Zhao, Wayne Xin and Nie, Jian-Yun and Wen, Ji-Rong},
  booktitle={Proceedings of the 2023 conference on empirical methods in natural language processing},
  pages={6449--6464},
  year={2023}
}

@article{geifman2018bias,
  title={Bias-reduced uncertainty estimation for deep neural classifiers},
  author={Geifman, Yonatan and Uziel, Guy and El-Yaniv, Ran},
  journal={arXiv preprint arXiv:1805.08206},
  year={2018}
}
\bibliographystyle{icml2026}

\newpage
\appendix
\onecolumn
\section{Full AURC Results}
\label{app:full-iid-aurc}

Table~\ref{tab:iid-aurc-main} in the main paper reports our method's AURC alongside the absolute improvement over the MSP baseline. For completeness, Table~\ref{tab:iid-aurc-full} below provides the full breakdown across all four methods we consider.

\begin{center}
\footnotesize
\setlength{\tabcolsep}{6pt}
\renewcommand{\arraystretch}{1.08}
\begin{tabular}{@{}ll
    S[table-format=2.2]
    S[table-format=2.2]
    S[table-format=2.2]
    S[table-format=2.2]@{}}
\toprule
& & \multicolumn{4}{c}{\textbf{AURC} $\downarrow$} \\
\cmidrule(lr){3-6}
\textbf{Model} & \textbf{Dataset} & {MSP} & {Ceiling} & {Ablation} & {Ours} \\
\midrule
Llama-3.2-3B   & MMLU      & 50.51 & 29.58 & 30.22 & 28.68 \\
               & CosmosQA  &  7.66 &  7.14 &  8.24 &  6.83 \\
               & HellaSwag & 33.60 & 32.86 & 36.20 & 31.99 \\
               & HaluDial  & 49.62 & 35.31 & 39.94 & 38.17 \\
               & HaluSum   & 50.92 & 23.12 & 34.97 & 34.38 \\
\cmidrule(l){2-6}
Llama-3.1-8B   & MMLU      & 26.47 & 24.64 & 25.62 & 22.65 \\
               & CosmosQA  &  3.85 &  3.73 &  4.46 &  3.58 \\
               & HellaSwag & 30.14 & 25.51 & 30.45 & 26.87 \\
               & HaluDial  & 29.07 & 19.93 & 23.98 & 22.48 \\
               & HaluSum   & 42.19 & 13.02 & 21.77 & 21.13 \\
\cmidrule(l){2-6}
Qwen2.5-7B     & MMLU      & 16.52 & 15.17 & 16.06 & 14.87 \\
               & CosmosQA  &  3.76 &  3.62 &  4.71 &  3.94 \\
               & HellaSwag &  8.66 &  5.96 &  7.16 &  6.59 \\
               & HaluDial  & 24.02 & 13.65 & 19.73 & 19.74 \\
               & HaluSum   & 30.32 & 10.60 & 20.94 & 20.67 \\
\cmidrule(l){2-6}
DeepSeek-7B    & MMLU      & 34.70 & 31.57 & 33.83 & 33.15 \\
               & CosmosQA  &  7.31 &  6.86 &  8.04 &  7.47 \\
               & HellaSwag & 25.97 & 18.42 & 20.94 & 20.65 \\
               & HaluDial  & 33.79 & 22.38 & 27.91 & 27.21 \\
               & HaluSum   & 43.59 & 14.84 & 29.85 & 29.30 \\
\cmidrule(l){2-6}
Qwen2.5-14B    & MMLU      & 11.09 & 10.74 & 11.54 & 10.86 \\
               & CosmosQA  &  2.40 &  1.91 &  2.56 &  2.24 \\
               & HellaSwag &  3.13 &  2.72 &  2.94 &  2.75 \\
               & HaluDial  & 15.99 &  9.74 & 12.58 & 12.05 \\
               & HaluSum   & 26.12 &  6.47 &  9.87 &  9.87 \\
\cmidrule(l){2-6}
DeepSeek-67B   & MMLU      & 11.63 & 11.51 & 12.35 & 11.83 \\
               & CosmosQA  &  1.49 &  1.53 &  1.78 &  1.50 \\
               & HellaSwag &  3.03 &  2.71 &  2.95 &  2.75 \\
               & HaluDial  & 14.73 &  7.81 &  9.69 &  9.38 \\
               & HaluSum   & 31.44 &  8.49 & 11.54 & 11.26 \\
\cmidrule(l){2-6}
Llama-3.3-70B  & MMLU      & 23.60 & 14.51 & 15.53 & 14.86 \\
               & CosmosQA  &  1.78 &  1.27 &  1.66 &  1.41 \\
               & HellaSwag &  4.71 &  3.00 &  3.30 &  3.00 \\
               & HaluDial  & 13.40 &  6.80 &  8.30 &  7.99 \\
               & HaluSum   & 30.27 &  7.18 & 10.15 & 10.10 \\
\cmidrule(l){2-6}
Qwen2-72B      & MMLU      &  8.31 &  5.78 &  6.30 &  6.10 \\
               & CosmosQA  &  1.69 &  1.11 &  1.39 &  1.27 \\
               & HellaSwag &  2.90 &  1.86 &  1.80 &  1.75 \\
               & HaluDial  & 11.74 &  6.26 &  7.36 &  7.29 \\
               & HaluSum   & 27.26 &  7.40 & 10.88 & 10.84 \\
\cmidrule(l){2-6}
Qwen2.5-72B    & MMLU      &  7.25 &  5.91 &  5.53 &  5.29 \\
               & CosmosQA  &  1.85 &  0.99 &  1.09 &  1.06 \\
               & HellaSwag &  4.57 &  1.76 &  1.82 &  1.82 \\
               & HaluDial  & 16.96 &  7.83 &  9.64 &  9.46 \\
               & HaluSum   & 25.56 &  5.74 &  8.81 &  8.82 \\
\bottomrule
\end{tabular}
\vspace{4pt}
\captionof{table}{Full AURC results ($\times 100$; lower is better). We compare our full probe (\textbf{ours}) against the MSP baseline, a raw-activations \textbf{ceiling} probe that operates on per-layer hidden states without capacity constraints, and a trajectory-only \textbf{ablation} that removes MSP and its interactions from the probe input.}
\label{tab:iid-aurc-full}
\end{center}

\section{Probe Hyperparameters}
\label{app:hyperparameters}

We fit the sparse linear probe of Section~\ref{sec:probe} using scikit-learn's \texttt{LogisticRegression} with the \texttt{saga} solver, class-balanced sample weighting, a convergence tolerance of $10^{-3}$, and a maximum of $8000$ iterations.  The $64$ configurations swept during model selection comprise $10$ pure-$\ell_1$ candidates with $C \in \{3 \!\times\! 10^{-4},\, 10^{-3},\, 3 \!\times\! 10^{-3},\, 10^{-2},\, 3 \!\times\! 10^{-2},\, 10^{-1},\, 3 \!\times\! 10^{-1},\, 1,\, 3,\, 10\}$, and $54$ elastic-net candidates over $C \in \{10^{-3},\, 3 \!\times\! 10^{-3},\, 10^{-2},\, 3 \!\times\! 10^{-2},\, 10^{-1},\, 3 \!\times\! 10^{-1},\, 1,\, 3,\, 10\}$ and $\ell_1$ mixing ratio $\rho \in \{0.10,\, 0.25,\, 0.50,\, 0.75,\, 0.90,\, 0.95\}$. All randomness in fitting and splitting is controlled by a fixed seed of $42$.

\end{document}